\documentclass[sigconf]{acmart}
\usepackage{array}
\usepackage{enumitem}
\usepackage{amsmath}
\usepackage{graphicx}
\usepackage{algorithm}
\usepackage{soul, color}
\usepackage{algpseudocode}
\usepackage{booktabs}
\usepackage{multirow}
\usepackage{float}
\usepackage{subcaption}
\usepackage{placeins}

\AtBeginDocument{%
  }

\setcopyright{acmlicensed}
\copyrightyear{2024}
\acmYear{2024}
\acmDOI{XXXXXXX.XXXXXXX}

\acmJournal{TOG}
\acmVolume{37}
\acmNumber{4}
\acmArticle{111}
\acmMonth{11}

\begin{document}


\title{FedPrism: Adaptive Personalized Federated Learning under Non-IID Data}

\author{Prakash Kumbhakar}
\affiliation{%
  \institution{IISER Bhopal}
  \city{Bhopal}
  \country{India}
}
\email{prakash22@iiserb.ac.in}

\author{Shrey Srivastava}
\affiliation{%
  \institution{IISER Bhopal}
  \city{Bhopal}
  \country{India}
}
\email{shrey22@iiserb.ac.in}

\author{Haroon R Lone}
\affiliation{%
  \institution{IISER Bhopal}
  \city{Bhopal}
  \country{India}
}
\email{haroon@iiserb.ac.in}

\begin{abstract}
Federated Learning (FL) suffers significant performance degradation in real-world deployments characterized by moderate to extreme statistical heterogeneity (non-IID client data). While global aggregation strategies promote broad generalization, they often fail to capture the diversity of local data distributions, leading to suboptimal personalization.

We address this problem with FedPrism, a framework that uses two main strategies. First, it uses a Prism Decomposition method that builds each client's model from three parts: a global foundation, a shared group part for similar clients, and a private part for unique local data. This allows the system to group similar users together automatically and adapt if their data changes. Second, we include a Dual-Stream design that runs a "general" model alongside a "local specialist." The system routes predictions between the general model and the local specialist based on the specialist’s confidence.

Through systematic experiments on non-IID data partitions, we demonstrate that FedPrism exceeds static aggregation and hard-clustering baselines, achieving significant accuracy gains under high heterogeneity. These results establish FedPrism as a robust and flexible solution for federated learning in heterogeneous environments, effectively balancing generalizable knowledge with adaptive personalization.  Our implementation is publicly available at \url{https://anonymous.4open.science/r/fedprism_ACM-7598}.
\end{abstract}

\maketitle


\section{Introduction}


Federated Learning (FL) models learn from decentralized data without compromising user privacy~\cite{li2021survey,li2020federated}. However, the foundational assumption of FL that a single one-size-fits-all model can serve a massive and diverse population is increasingly proving to be its greatest weakness \cite{li2020federated}. In real-world deployments, data is rarely Independent and Identically Distributed (IID) \cite{kairouz2021advances}. Instead, clients operate in statistical silos where differences in geography or usage habits create extreme non-IID data distributions (Non-IID)~\cite{li2020federated}.

The tension in FL lies in the Personalization Paradox, where a global model provides robustness through shared knowledge while local training provides precision through specialization. We argue that a client identity is rarely static or pure \cite{mansour2020three}. A robust FL system must acknowledge that clients are mixtures of different latent patterns \cite{mansour2020three}. A user might share commonalities with one group in certain tasks but remain unique in others. There is a critical need for a framework that not only personalizes but also decomposes the learning process. By dynamically balancing shared global foundations, group-level expertise, and private local nuances, we can ensure no client is left behind by the average.

Existing approaches for handling statistical heterogeneity (i.e., Non-IID data) in federated learning largely fall into three paradigms: global regularization, personalization via parameter separation, and clustered federated learning. Regularization-based methods such as FedProx \cite{li2020federated} and SCAFFOLD \cite{karimireddy2020scaffold} improve convergence stability under non-IID data, yet they still optimize a single global objective. Consequently, they lack the flexibility to represent fundamentally diverse client distributions. Personalized methods including FedPer \cite{arivazhagan2019federated}, pFedMe \cite{t2020personalized}, and Ditto \cite{li2021ditto} decouple shared and local parameters to enable client-specific adaptation. However, these approaches do not explicitly model structured relationships among groups of similar clients, limiting their ability to exploit latent distributional structure. Clustered federated learning attempts to address this limitation by grouping clients with similar data distributions. Nevertheless, existing clustered methods exhibit important design constraints. Static clustering approaches such as FedClust~\cite{islam2024fedclust} assume fixed client-group assignments, while IFCA~\cite{ghosh2020efficient} employs hard assignments that restrict each client to a single latent distribution \cite{lu2024federated}. In realistic federated environments, client data is often heterogeneous and may follow distinct but related distributions, making the assumption of a single shared model or rigid cluster assignment overly restrictive~\cite{sattler2020clustered}. Moreover, in its standard formulation, IFCA transmits all cluster models to every client at each communication round, resulting in substantial bandwidth overhead. Soft-clustering approaches such as FedAMP \cite{huang2021personalized} aim to relax hard assignments, but they introduce significant computational complexity through pairwise client interactions and attention-based aggregation, which limits scalability in large or resource-constrained federated systems.

While FedEM \cite{marfoq2021federated} models clients as mixtures of latent distributions, its mixture assignments remain static and it does not incorporate dynamic re-clustering or inference-time confidence-aware routing. Consequently, there remains no unified framework that combines dynamic clustering with computationally efficient soft multi-assignment to enable scalable, high-fidelity personalization under evolving and highly heterogeneous data distributions.

We propose \emph{FedPrism} (Federated Personalized Relevance-based Intelligent Soft-assignment Model), a framework that addresses rigidity in federated optimization. 
FedPrism introduces dynamic clustering that updates client similarity during training to handle non stationary data \cite{kairouz2021advances, yang2023impact}. Unlike hard partitioning \cite{ghosh2020efficient}, we employ \emph{soft assignment}, modeling each client as a weighted combination of latent clusters. This design connects to mixture of experts models \cite{jacobs1991adaptive, shazeer2017outrageously} and federated multi task learning \cite{smith2017federated}, enabling partial knowledge sharing while reducing negative transfer. Moreover, we introduce \emph{Prism Decomposition}, where each client model consists of Global, Cluster, and Personal components; and design a \emph{Dual Stream architecture} with a collaborative backbone and an independent local expert. Inspired by conditional computation and expert routing \cite{shazeer2017outrageously} and uncertainty modeling \cite{kendall2017uncertainties}, a confidence aware routing mechanism balances generalization and specialization, mitigating performance degradation under heterogeneous client distributions \cite{li2021ditto}. To support reproducibility, we release the complete
implementation of \emph{FedPrism} at \url{https://anonymous.4open.science/r/fedprism_ACM-7598}.

We evaluate FedPrism on multiple benchmarks under controlled non IID settings \cite{zhao2018federated, li2020federated}. We compare against strong baselines spanning local training, global aggregation, clustered federated learning, and personalized collaboration, including FedAvg \cite{mcmahan2017communication}, IFCA \cite{ghosh2020efficient}, FedAMP \cite{huang2021personalized}, and Clustered Federated Learning \cite{sattler2020clustered}.  Results show consistent gains under moderate and extreme heterogeneity. Ablation studies confirm the individual contributions of dynamic clustering, soft assignment, and structured decomposition, establishing FedPrism as a robust and adaptive solution for personalized federated learning.

\section{Related Work}
The challenge of non-IID data in federated learning has motivated extensive research in personalized and clustered FL approaches \cite{huang2021personalized, yan2023clustered, tan2022towards}. We review recent work in clustered and personalized federated learning, highlighting the algorithmic limitations that FedPrism addresses.

\textbf{Clustered Federated Learning:} To address non-IID data fundamentally, clustered FL methods group clients with similar distributions. Islam et al.~\cite{islam2024fedclust} proposed FedClust, which groups clients based on the similarity of their model's final layer weights. However, this clustering is performed one-shot at the beginning and remains static, making it unable to adapt to concept drift or correct suboptimal initial clustering. Similar rigidity appears in hierarchical clustering-based federated learning approaches \cite{briggs2020federated}, which organize clients at multiple
levels but still rely on hard cluster membership. Ghosh et al.~\cite{ghosh2020efficient} introduced IFCA (Iterative Federated Clustering Algorithm), a foundational dynamic clustering approach where the server maintains $k$ cluster models. While IFCA addresses the static clustering limitation, it employs hard-assignment, forcing clients to pick exactly one cluster. This is suboptimal for clients with mixed or hybrid data distributions and limits the model's ability to capture complex data patterns. Sattler et al.~\cite{sattler2020clustered} proposed Clustered Federated Learning (CFL), which clusters clients after FL convergence based on similarities in model updates without modifying the standard FL protocol or predefining the number of clusters. However, since clustering is performed only as a post-processing step with hard partitioning, it lacks dynamic adaptation and flexible soft personalization for mixed data distributions.

\textbf{Personalized Federated Learning:} Beyond clustering, personalization methods aim to tailor models to individual clients. Huang et al.~\cite{huang2021personalized} proposed FedAMP, a method that enables soft clustering through an attentive message-passing mechanism on the server, allowing clients to communicate more with similar peers. While this approach achieves personalization, the server-side attentive graph computation is highly complex and adds significant computational overhead, making it difficult to tune and scale compared to simpler aggregation methods. Other approaches rely on meta-learning or regularization-based personalization,
which can mitigate local drift but often fail to facilitate positive transfer
between distinct yet related groups of clients. Li et al.~\cite{li2020federated}
proposed FedProx, a proximal extension of FedAvg that stabilizes local updates
under statistical and systems heterogeneity via a device-level regularization term.
Similarly, Dinh et al.~\cite{t2020personalized} introduced pFedMe, which personalizes
client models by constraining local optimization around a shared global solution
using Moreau envelopes. While effective for stabilizing training, such
regularization-based methods do not explicitly model multi-modal client similarity
or enable soft knowledge sharing across multiple clusters.
 In FedRep~\cite{collins2021exploiting}, a shared global feature representation is collaboratively learned across clients while keeping client-specific prediction heads local, enabling efficient personalization by isolating heterogeneity to the output layers and reducing the dimensionality of local optimization. Generally, multi-task FL frameworks such as MOCHA
\cite{smith2017federated} explicitly model task relationships across clients,
but incur substantial optimization and communication overhead, limiting their
scalability in large and dynamic federated settings.

\begin{figure*}[t]
    \centering
    \includegraphics[width=0.9\textwidth]{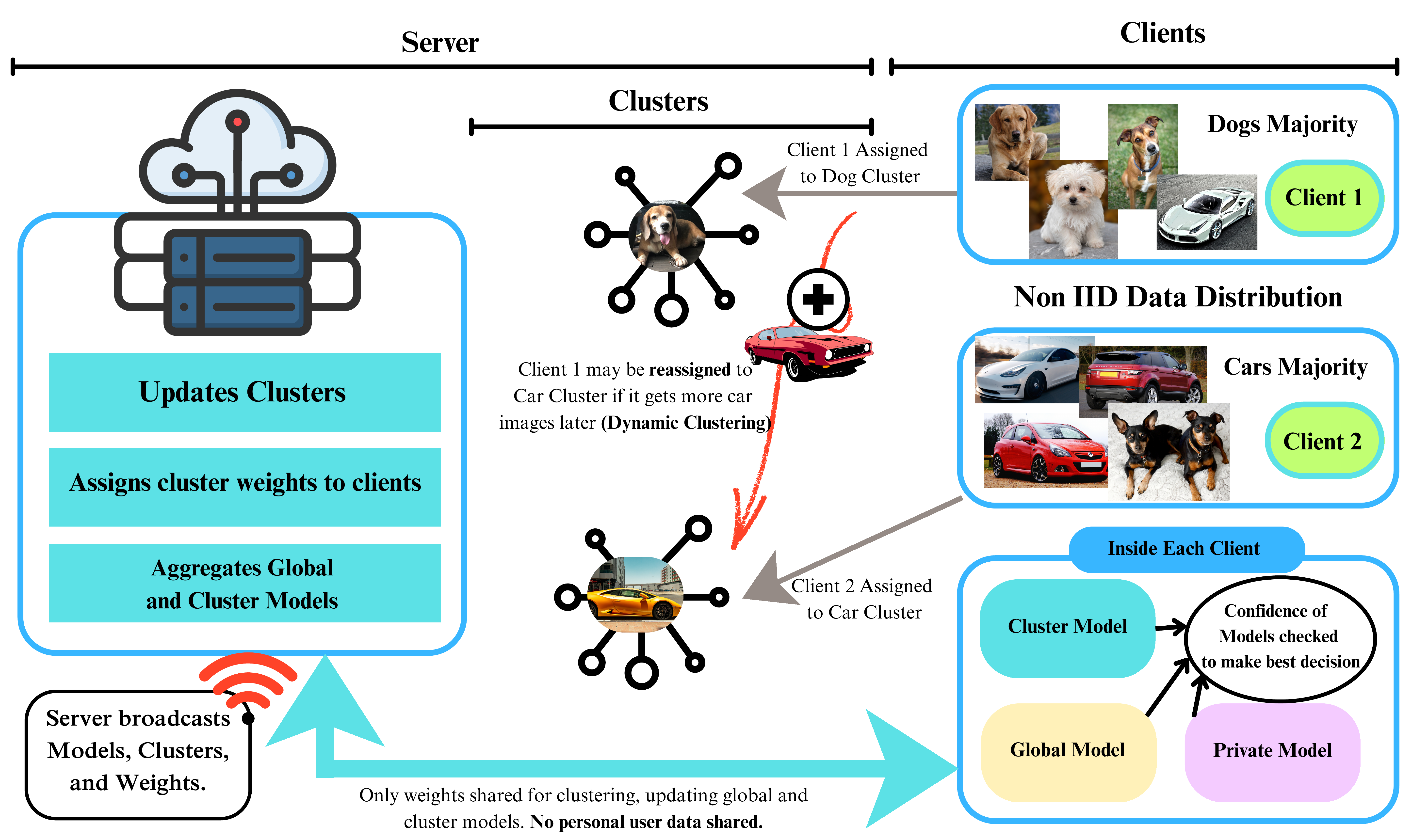}
    \caption{Overview of the FedPrism framework illustrating dynamic clustering, soft assignment, and the Global–Cluster–Personal (GCP) model decomposition within the federated learning pipeline.}
    \label{fig:fed_prism_overview}
\end{figure*}


\section{Methodology}
\label{sec:methodology}

In this section, we present the \textbf{FedPrism} framework (see Figure \ref{fig:fed_prism_overview}), a novel federated learning approach designed to address the "one-size-fits-all" limitation of traditional aggregation. We introduce two key ideas: splitting the model into three parts (Prism Decomposition) and using two separate models during inference (Dual-Stream Architecture). The complete FedPrism training workflow is formally described in Algorithms~\ref{alg:fedprism_client} and~\ref{alg:fedprism_server}.

\subsection{Prism Decomposition}

The main idea of FedPrism is to split the model weights into three separate components. Instead of just having one global model, each client's model is built by adding three parts together. This allows the model to learn general patterns, group-specific patterns, and client-specific patterns all at the same time.

Mathematically, for a client $i$, the model weights $w_i$ are defined as:

\begin{equation}
w_i = \underbrace{\alpha_i w_G}_{\text{Global (G)}} + \underbrace{\beta \sum_{k=1}^K \pi_{i,k} C_k}_{\text{Cluster (C)}} + \underbrace{\gamma_i P_i}_{\text{Private (P)}}
\end{equation}

Here is what each part does:

\begin{itemize}
    \item \textbf{Global Component ($w_G$):} This is the standard shared model. It is trained by averaging updates from all clients. It learns general features that are useful for everyone, such as basic shapes or edges in images. It acts as a stable foundation for the model.
    
    \item \textbf{Cluster Component ($\sum \pi_{i,k} C_k$):} This part helps clients that are similar to each other. The server maintains $K$ different "cluster models" ($C_1$ to $C_K$). Each client calculates how similar it is to these clusters and assigns itself a weight $\pi_{i,k}$ for each one. For example, if a client has mostly animal images, it will give a high weight to the "animal cluster" model. This allows the client to use knowledge from similar peers without needing to be in a strict group.
    
    \item \textbf{Private Component ($P_i$):} This part is specific to the client. It is trained only on the client's local data and is never shared with the server. It learns the unique details of the local dataset that the global or cluster models might miss.
\end{itemize}

The parameters $\alpha_i$, $\beta$, and $\gamma_i$ control the relative contribution of each component. 
The value of $\alpha_i$ is updated automatically during training to balance the influence of the global backbone and the client-specific components.

\subsection{Dynamic Prototype-Based Clustering}

To make the Cluster Component work, we need to know which clients are similar. FedPrism does this by grouping clients based on their model's "prototypes" (a summary of what the model has learned).

The process happens in three steps every round:
\begin{enumerate}
    \item \textbf{Step 1: Extract Prototype.} Each client looks at the last layer of its model (the classifier). We treat the weights of this layer as a "feature prototype" that represents the client's data distribution.
    \item \textbf{Step 2: Server Clustering.} The server collects these prototypes from all active clients. It runs the K-Means algorithm \cite{ikotun2023k} to find $K$ centers (centroids) that represent the main groups of clients.
    \item \textbf{Step 3: Calculate Similarity.} The client compares its own prototype to these $K$ centers using cosine similarity. If a client is very close to Center 1, it assigns a high weight ($\pi_{i,1}$) to Cluster Model 1. This creates a "soft" assignment, meaning a client can belong to multiple clusters to varying degrees.
\end{enumerate}

This method allows the system to adapt over time. If a client's data helps it learn new features, its prototype will change, and it will automatically shift to a more relevant cluster.

\subsection{Dual-Stream Robust Architecture}

Even with the Prism Decomposition, a single model can struggle if the data is extremely different across clients (e.g., if Client A has only dogs and Client B has only cars). To fix this, FedPrism uses a \textbf{Dual-Stream Architecture}. This means we actually maintain and train two separate models for each client:

\begin{enumerate}
    \item \textbf{The Global Backbone:} This is the collaborative model described above ($G+C+P$). It is good at generalization but might be less accurate on specific local classes.
    \item \textbf{The Local Expert:} This is a completely independent model trained \textbf{only} on the client's local data. It is a specialist for the local tasks but cannot generalize to unseen data.
\end{enumerate}

\subsubsection{Confidence-Aware Inference Routing}
When the system makes a prediction for a new input x, it must determine whether to rely on the generalized backbone or the specialized local expert.

We use the confidence of the Local Expert to make this decision. We assume that if the Local Expert sees an image similar to its training data, it will be very confident. If it sees something new, it will be uncertain.

To control how "sharp" or "soft" this decision is, we use a temperature parameter $T$. We scale the expert's output by $T$ before calculating the probability.

The routing weight $\lambda(x)$ is the maximum probability output:
\begin{equation}
    \lambda(x) = \max\left(\text{Softmax}\left(\frac{\text{Expert}(x)}{T}\right)\right)
\end{equation}
The final prediction $y_{pred}$ is a weighted average of both models:
\begin{equation}
    y_{pred} = \lambda(x) \cdot \text{Expert}(x) + (1 - \lambda(x)) \cdot \text{Backbone}(x)
\end{equation}

\begin{itemize}
    \item If the Local Expert is \textbf{confident} (high $\lambda$), the prediction relies primarily on the Expert.
    \item If the Local Expert is \textbf{uncertain} (low $\lambda$), the prediction relies primarily on the Backbone.
\end{itemize}

This ensures we get the best of both worlds: high accuracy on familiar data (from the Expert) and robustness on new data (from the Backbone).

\begin{algorithm}[t]
\caption{Client-Side: FedPrism Dual Robust (Specialist/Generalist Updates)}
\label{alg:fedprism_client}
\begin{algorithmic}[1]
\State \textbf{Input:} Local dataset $\mathcal{D}_i$
\State \textbf{Receive from Server:} Global $G^{(t)}$, Clusters $\{C_k^{(t)}\}_{k=1}^K$, Weights $\{w_{ik}^{(t)}\}$, $\alpha_i^{(t)}$
\State \textbf{Maintain Locally:} 
    \par\hskip\algorithmicindent $\bullet$ Private Component $P_i^{(t)}$
    \par\hskip\algorithmicindent $\bullet$ Local Specialist $L_i^{(t)}$
\State \textbf{Hyperparameters:} $\eta$ (Learning Rate), $\beta$ (Cluster Coeff)

\Procedure{ClientUpdate}{}
    \State \textit{// Path A: Train Independent Local Specialist}
    \State $L_i^{(t+1)} \leftarrow L_i^{(t)} - \eta \nabla \mathcal{L}(L_i^{(t)}; \mathcal{D}_i)$

    \State \textit{// Path B: Train Shared Generalist}
    \State \textit{Step 1: Update Private Component}
    \State $P_i^{(t+1)} \leftarrow P_i^{(t)} - \eta \nabla \mathcal{L}(P_i^{(t)}; \mathcal{D}_i)$
    
    \State \textit{Step 2: Construct \& Train Mixed Prismatic Model}
    \State $\gamma_i \leftarrow 1 - \alpha_i^{(t)} - \beta$
    \State $W_{\text{Backbone}}^{(t)} \leftarrow \alpha_i^{(t)} G^{(t)} + \beta \sum_{k=1}^K w_{ik}^{(t)} C_k^{(t)} + \gamma_i P_i^{(t+1)}$
    \State $W_{\text{Backbone}}' \leftarrow W_{\text{Backbone}}^{(t)} - \eta \nabla \mathcal{L}(W_{\text{Backbone}}^{(t)}; \mathcal{D}_i)$
    
    \State \textit{Step 3: Feature Extraction}
    \State $h_i^{(t)} \leftarrow \text{ExtractFeatures}(W_{\text{Backbone}}')$

    \State \Return $W_{\text{Backbone}}', h_i^{(t)}$
\EndProcedure

\Procedure{Inference}{$x$}
    \State \textit{// Dual-Stream Routing (Test Time)}
    \State $z_L \leftarrow L_i^{(t+1)}(x)$
    \State $p_L \leftarrow \text{Softmax}(z_L / T)$
    \State $c \leftarrow \max(p_L)$
    \Statex \hspace{\algorithmicindent} \textit{// Confidence Score}
    
    \State $z_G \leftarrow W_{\text{Backbone}}(x)$
    \State $z_{\text{final}} \leftarrow c \cdot z_L + (1-c) \cdot z_G$
    \State \Return $\arg\max(z_{\text{final}})$
\EndProcedure

\end{algorithmic}
\end{algorithm}

\begin{algorithm}[t]
\caption{Server-Side: FedPrism Aggregation \& Clustering}
\label{alg:fedprism_server}
\begin{algorithmic}[1]
\State \textbf{Initialize:} $G^{(0)}$, $\{C_k^{(0)}\}$, $\{w_{ik}^{(0)}\}$, $\{\alpha_i^{(0)}\}$
\State \textbf{Hyperparameters:} $\eta_{\text{cluster}}$, $\eta_{\alpha}$

\For{round $t = 0,1,\dots,T$}
    \State Sample active clients $\mathcal{S}_t \subseteq \{1,\dots,N\}$

    \For{each client $i \in \mathcal{S}_t$ \textbf{in parallel}}
        \State Send $G^{(t)}, \{C_k^{(t)}\}, w_{ik}^{(t)}, \alpha_i^{(t)}$
        \State Receive $W_{\text{Backbone},i}', h_i^{(t)}$
    \EndFor

    \State \textit{// A. Global Aggregation}
    \State $G^{(t+1)} \leftarrow \sum_{i \in \mathcal{S}_t} \frac{n_i}{N_t} W_{\text{Backbone},i}'$

    \State \textit{// B. Cluster Updates (Moving Average)}
    \For{$k = 1$ to $K$}
    \State \textit{// Soft cluster aggregation using assignment weights}
    \State $Z_k \leftarrow \sum_{i \in \mathcal{S}_t} w_{ik}^{(t)}$
    \If{$Z_k > 0$}
        \State $\bar{W}_k \leftarrow 
        \frac{1}{Z_k}
        \sum_{i \in \mathcal{S}_t}
        w_{ik}^{(t)} \cdot W_{\text{Backbone},i}'$
        \State $C_k^{(t+1)} \leftarrow 
        C_k^{(t)} + \eta_{\text{cluster}}
        (\bar{W}_k - C_k^{(t)})$
    \EndIf
\EndFor

    \State \textit{// C. Update Cluster Assignments}
    \If{$t > \text{Warmup}$}
        \State Update centroids $\{\mu_k^{(t+1)}\}$ via K-Means on received $\{h_i^{(t)}\}$
        \For{each client $i \in \mathcal{S}_t$}
            \State $w_{ik}^{(t+1)} \propto \exp\!\left(\text{sim}(h_i^{(t)}, \mu_k^{(t+1)}) / \tau\right)$
        \EndFor
    \EndIf

    \State \textit{// D. Update Mixing Weights}
    \For{each client $i \in \mathcal{S}_t$}
        \State $\Delta \leftarrow \|W_{\text{Backbone},i}' - C_{k}^{(t)}\| - \|W_{\text{Backbone},i}' - G^{(t)}\|$
        \State $\alpha_i^{(t+1)} \leftarrow \text{Clip}(\alpha_i^{(t)} + \eta_{\alpha} \Delta, [0, 1])$
    \EndFor

\EndFor
\end{algorithmic}
\end{algorithm}

\section{Experimental Setup}
This section provides a comprehensive description of the datasets, non-IID data partitioning strategy, baseline algorithms, model architecture, and implementation specifications. The goal is to rigorously evaluate the effectiveness of the proposed \textbf{FedPrism} framework in addressing data heterogeneity compared to state-of-the-art Federated Learning methods.

\subsection{Datasets}

We evaluate the performance of FedPrism using four standard image classification benchmarks widely used in FL research:
\begin{itemize}
    \item \textbf{CIFAR-10}~\cite{krizhevsky2012imagenet}: Consists of 60,000 $32\times32$ color images relative to 10 classes (airplane, automobile, bird, etc.). 
    \item \textbf{CIFAR-100}~\cite{krizhevsky2012imagenet}: Similar to CIFAR-10 but contains 100 classes (600 images per class), serving as a benchmark for fine-grained classification tasks under federated settings.
    \item \textbf{SVHN (Street View House Numbers)}~\cite{goodfellow2013multi}: A real-world digit recognition dataset obtained from house numbers in Google Street View images.
    \item \textbf{Fashion-MNIST (FMNIST)}~\cite{xiao2017fashion}: Comprises 60,000 $28\times28$ grayscale training images and 10,000 test images across 10 fashion categories (t-shirt, trouser, pullover, etc.).
\end{itemize}

\subsection{Data Partitioning (Non-IID Simulation)}

To simulate statistical heterogeneity, we partition the dataset across $N=100$ clients using two widely adopted non-IID strategies: Dirichlet-based partitioning and pathological (shard-based) partitioning. 

Under the Dirichlet distribution-based approach, we generate label imbalance by sampling, for each client $i$, a class proportion vector $\mathbf{p}_i \sim Dir(\alpha \mathbf{1})$, where $\mathbf{p}_{i,k}$ denotes the probability of class $k$ in client $i$'s local dataset. Training samples are allocated according to $\mathbf{p}_i$, allowing controlled variation in non-IID severity through the concentration parameter $\alpha$. We consider three heterogeneity regimes. When $\alpha=0.5$, the setting corresponds to moderate non-IID. For $\alpha=0.3$, we obtain a high non-IID scenario with reduced class overlap across clients. Finally, $\alpha=0.1$ yields an extreme non-IID regime, where clients typically contain one or two dominant classes, creating severe statistical divergence and strong client drift.

To model an even more extreme and structured label-skew scenario, we additionally adopt the classical pathological partitioning strategy. In this setting, the dataset is first sorted by class labels and divided into multiple shards. Each client is assigned a limited number of shards, resulting in local datasets that contain samples from only a small subset of classes. This construction enforces strong class isolation and represents a highly heterogeneous federated environment. The pathological partitioning setting is implemented for CIFAR-10, CIFAR-100, and SVHN, while Dirichlet-based partitioning is applied consistently across all datasets for systematic comparison.

\begin{figure}[t]

    \centering
    \includegraphics[width=\linewidth]{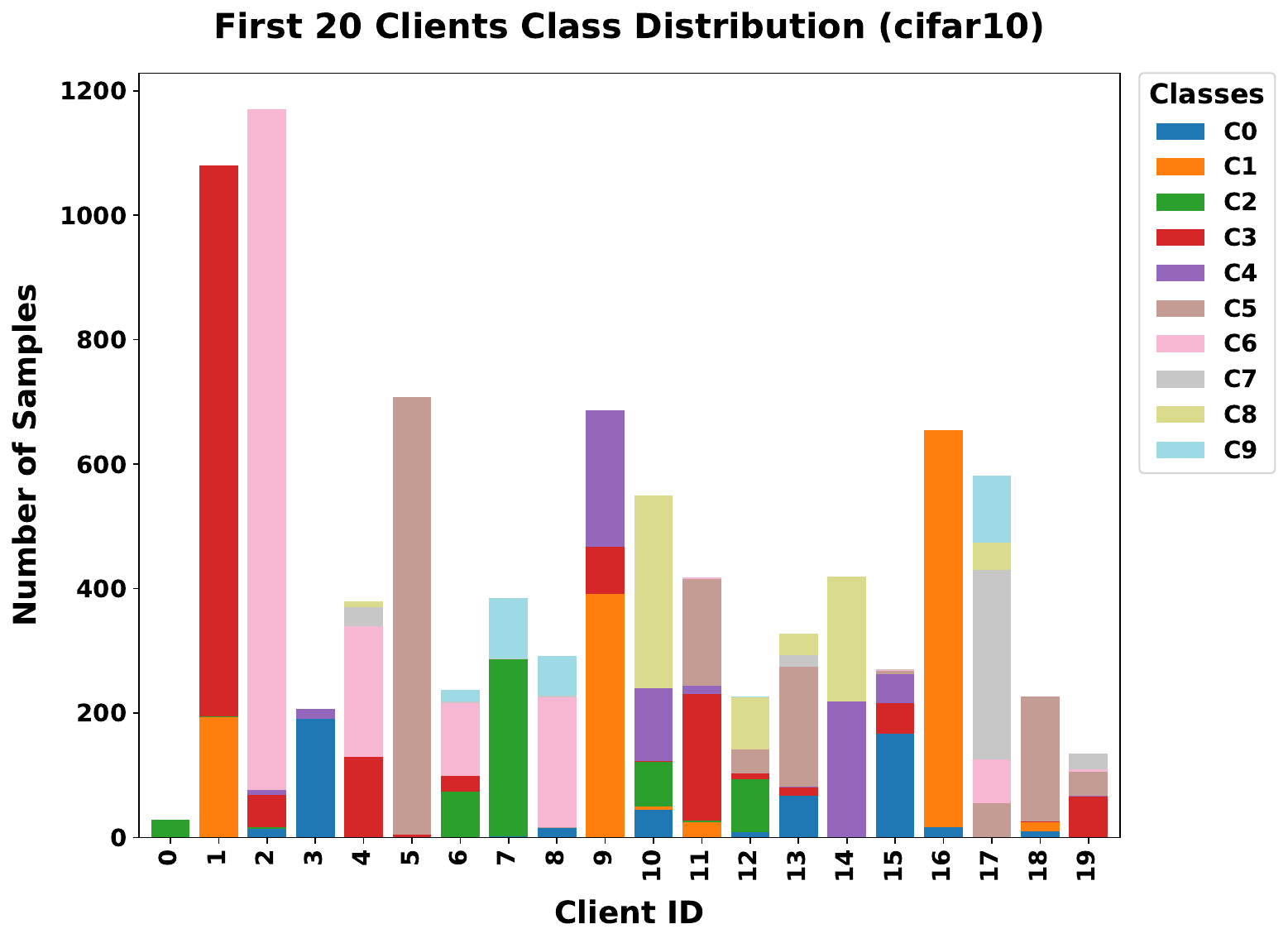}
    \caption{Class distribution across the first 20 clients under Dirichlet partitioning with $\alpha=0.1$ on CIFAR-10.}
    \label{fig:cifar10_dirichlet_01}
\end{figure}

\begin{figure}[t]

    \centering
    \includegraphics[width=\linewidth]{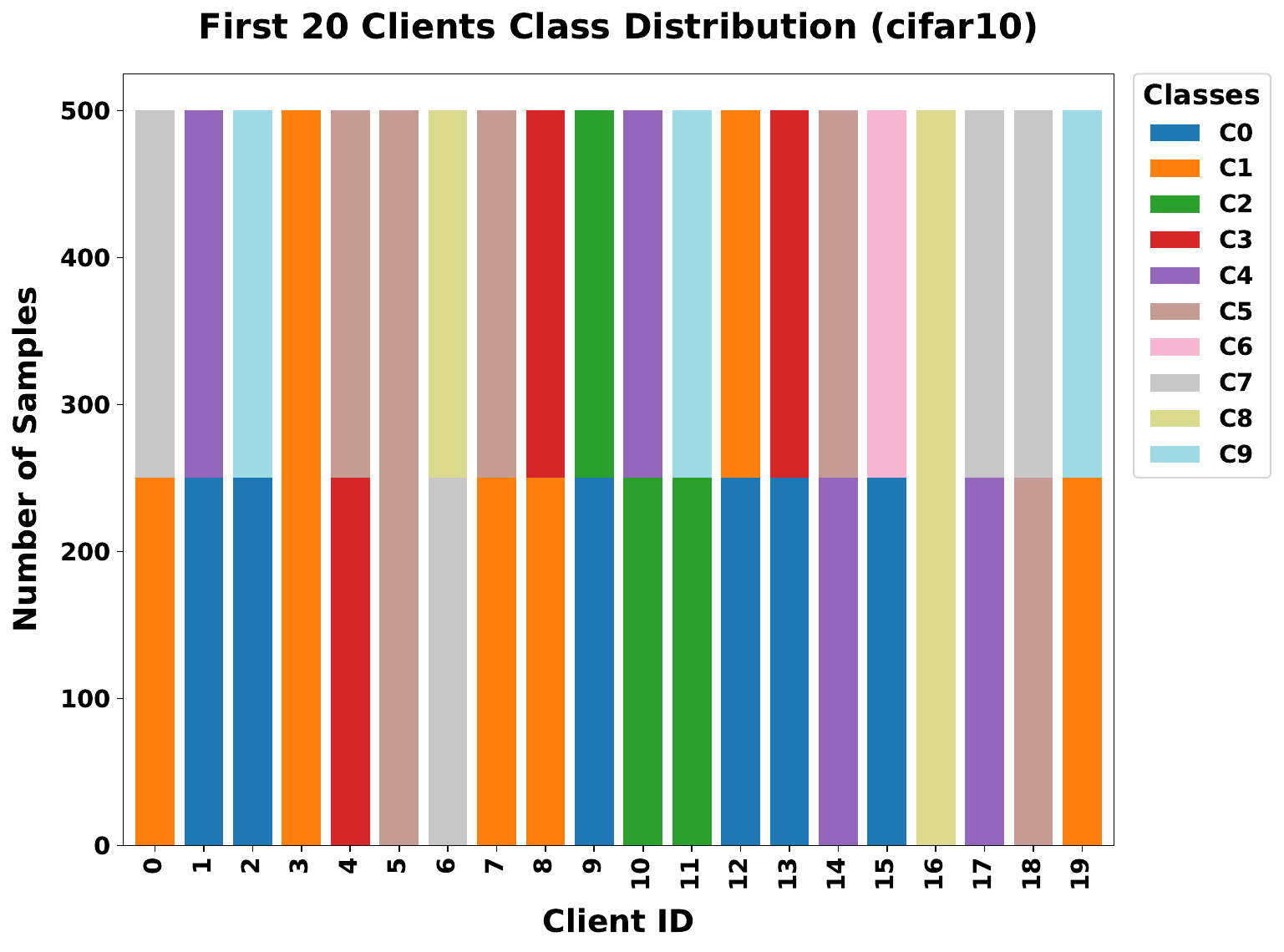}
    \caption{Class distribution across the first 20 clients under pathological (shard-based) partitioning on CIFAR-10. Each client contains samples from a limited subset of classes, demonstrating extreme label skew.}
    \label{fig:cifar10_pathological}
\end{figure}

To explicitly illustrate the induced heterogeneity, Figure~\ref{fig:cifar10_dirichlet_01} presents the class distribution of the first 20 clients under the Dirichlet partitioning with $\alpha=0.1$ for CIFAR-10. Figure~\ref{fig:cifar10_pathological} shows the corresponding distribution for CIFAR-10 under the pathological partitioning strategy, also for the first 20 clients. These visualizations clearly demonstrate the concentration of labels and the reduced class overlap across clients in both extreme non-IID settings. 
For other datasets,the distribution plots are provided in Appendix \ref{appendix:data_distribution}

For consistency and to avoid distribution mismatch during evaluation, the same sampled client-specific distribution $\mathbf{p}_i$ is applied to both the training and test splits.

\subsection{Model Architecture}
For all datasets, we employ a LeNet-5 \cite{lecun2002gradient} Convolutional Neural Network (CNN). This architecture is widely used in federated learning research \cite{theologitis2024communication} \cite{lutz2025optimizing} due to its computational efficiency, small parameter size, and strong performance on standard benchmark datasets, which facilitates fair comparison with prior work. The network was adapted to handle both $32\times32$ (CIFAR, SVHN) and $28\times28$ (FMNIST) inputs (via padding) as follows:
\begin{enumerate}
    \item \textbf{Conv1}: $5\times5$ convolution, 6 filters, ReLU activation.
    \item \textbf{MaxPool1}: $2\times2$ max pooling.
    \item \textbf{Conv2}: $5\times5$ convolution, 16 filters, ReLU activation.
    \item \textbf{MaxPool2}: $2\times2$ max pooling.
    \item \textbf{FC1}: Fully connected layer (400 $\rightarrow$ 120), ReLU.
    \item \textbf{FC2}: Fully connected layer (120 $\rightarrow$ 84), ReLU.
    \item \textbf{Output}: Fully connected layer (84 $\rightarrow$ Num\_Classes).
\end{enumerate}

\begin{table*}[t]
\centering
\caption{Performance Comparison under Extreme Heterogeneity (Dirichlet $\alpha=0.1$). All values are accuracy (\%). Results are reported as mean $\pm$ standard deviation (0.5--0.9\%). FedPrism  consistently achieves superior local personalization while maintaining competitive global performance.}
\label{tab:dir_0.1}
\begin{tabular}{l|cc|cc|cc|cc}
\hline
\multirow{2}{*}{\textbf{Algorithm}} 
& \multicolumn{2}{c|}{\textbf{SVHN (Acc \%)}} 
& \multicolumn{2}{c|}{\textbf{FMNIST (Acc \%)}} 
& \multicolumn{2}{c|}{\textbf{CIFAR-10 (Acc \%)}} 
& \multicolumn{2}{c}{\textbf{CIFAR-100 (Acc \%)}} \\ \cline{2-9} 
 & \textit{Glob} & \textit{Loc} 
 & \textit{Glob} & \textit{Loc} 
 & \textit{Glob} & \textit{Loc} 
 & \textit{Glob} & \textit{Loc} \\ \hline
\textbf{FedPrism (Ours)} & 73.67 & \textbf{87.89} & 81.93 & \textbf{95.66} & 38.69 & \textbf{81.63} & \textbf{14.33} & \textbf{39.91} \\ \hline
FedAvg & \textbf{83.11} & 82.16 & 84.35 & 85.01 & 36.83 & 36.11 & 13.46 & 13.48 \\
IFCA & 51.27 & 83.82 & 35.52 & 87.05 & 18.41 & 55.28 & 6.81 & 15.50 \\
FedClust & 44.54 & 36.51 & 10.00 & 58.11 & 10.56 & 17.77 & 4.96 & 8.00 \\
FedAMP & 11.07 & 51.95 & 10.36 & 55.58 & 10.00 & 56.24 & 1.19 & 39.01 \\
FedCFL & 6.38 & 17.72 & 10.00 & 44.86 & 10.00 & 11.73 & 1.06 & 3.89 \\
Local & 23.45 & 84.93 & 30.51 & 95.05 & 16.01 & 79.84 & 4.44 & 38.05 \\ \hline
\end{tabular}%
\end{table*}

\subsection{Baselines}
We benchmark \textbf{FedPrism} against the following approaches:
\begin{enumerate}
    \item \textbf{Local Training}: Each client trains the model solely on its own local dataset without any communication or parameter sharing with other clients. This setup serves as a baseline to evaluate the gains achieved through federated collaboration.
    \item \textbf{FedAvg} \cite{sun2022decentralized}: The canonical FL algorithm. The server aggregates client updates via weighted averaging: $\mathbf{w}^{t+1} = \sum_{k \in S_t} \frac{n_k}{n} \mathbf{w}_k^{t+1}$.
    \item \textbf{FedClust} \cite{islam2024fedclust}: A clustered FL method that groups clients based on the cosine similarity of their final layer weights (or gradients). The server forms $C$ clusters using K-Means and aggregates models within each cluster. We implement a specific stability fix using the Hungarian algorithm \cite{kuhn1955hungarian} to match cluster identities across rounds.
    \item \textbf{IFCA (Iterative Federated Clustering Algorithm)} \cite{ghosh2020efficient}: The server maintains $K$ global models. In each round, selected clients evaluate all $K$ models on their local data, select the one with the lowest loss, update it, and send it back. The server aggregates updates for each model index $k$ separately.
    \item \textbf{FedAMP (Federated Attentive Message Passing)} \cite{huang2021personalized}: 
A personalized FL method that replaces global averaging with attention-based aggregation. The server computes pairwise model similarity and assigns attention weights $a_{ij}^t$, enabling each client to receive a weighted combination of similar clients’ models:
\[
\mathbf{w}_i^{t+1} = \sum_{j \in S_t} a_{ij}^t \mathbf{w}_j^t.
\]
This promotes adaptive collaboration while preserving client-specific personalization.

\item \textbf{Clustered Federated Learning (CFL)} \cite{sattler2020clustered}: A federated multitask learning approach that clusters clients based on cosine similarity of their model updates at convergence. It detects data heterogeneity and recursively splits clients without modifying the standard FL protocol. CFL does not require knowing the number of clusters in advance and supports nonconvex models. 

\end{enumerate}

\subsection{Implementation Details}
All algorithms are implemented in PyTorch. Training is conducted synchronously for $R=100$ global rounds, with a fraction $C=0.1$ of clients participating per round. Each selected client performs $E=10$ local epochs using SGD with learning rate $\eta=0.01$, momentum $0.9$, and batch size $B=32$. For FedPrism, we initialize $K=5$ clusters and apply K-Means clustering every 10 rounds to support dynamic personalization. Results are averaged over multiple random seeds for robustness. 
 In each round, we report \textbf{Global Accuracy} on the full official IID test set to measure generalization, and \textbf{Local Accuracy} on client-specific non-IID test subsets constructed using the same partitioning strategy (Dirichlet or pathological) as training, to assess personalization performance.

 The experiments were conducted on a workstation equipped with an Intel Core i7-14700 processor (14th Gen) featuring 20 physical cores and 28 threads with a maximum clock frequency of 5.4 GHz, alongside 32 GB of system RAM. The system includes an NVIDIA GeForce RTX 4070 GPU with 16 GB of GDDR6 memory, running CUDA 13.0. All experiments were implemented in Python 3.13.11.

\section{Results}
\label{sec:results}

We present the performance of \textbf{FedPrism} across varying degrees of data heterogeneity, focusing on its ability to balance global knowledge aggregation with local personalization. 

\subsection{Performance under Extreme Heterogeneity ($\alpha=0.1$)}
Table~\ref{tab:dir_0.1} details the performance under the most challenging Dirichlet setting ($\alpha=0.1$), where client data distributions are highly skewed. In this regiment, the superiority of FedPrism's personalization mechanism becomes evident. On complex datasets like CIFAR-100, standard global aggregation methods such as FedAvg fail to capture the diverse local distributions, resulting in a low local accuracy of 13.48\%. Similarly, FedClust struggles to find meaningful clusters amidst the noise, achieving only 8.00\%. 

In sharp contrast, \textbf{FedPrism} achieves a Local Accuracy of \textbf{39.91\%}, practically triple the performance of the strongest global baseline. This output suggests that FedPrism's dual-stream architecture successfully decouples the global representation learning from local classifier specialization. The global stream learns generalizable features that arguably prevent overfitting, while the local stream fine-tunes these features for the specific classes present on the client. 

This trend of personalization dominance holds true even for simpler datasets. On FMNIST, FedPrism reaches near-perfect personalization with \textbf{95.66\%} local accuracy, significantly outpacing FedAvg's 85.01\%. Comparing this to the Local-only baseline (95.05\%) reveals a crucial insight: unlike FedAvg, which can sometimes degrade performance below local training due to "negative transfer" from dissimilar clients, FedPrism successfully filters interfering signals, ensuring that participation in the federation is strictly additive to performance.

Under moderate ($\alpha = 0.3$) and low ($\alpha = 0.5$) heterogeneity settings, FedPrism consistently demonstrates strong personalization benefits while maintaining competitive global performance across datasets. In the moderately heterogeneous case, it achieves notable gains in local accuracy, particularly on more complex tasks such as CIFAR-100, highlighting the importance of personalization under distributional diversity. Even in the lower heterogeneity setting, where data becomes more IID and personalization advantages typically diminish, FedPrism remains competitive and continues to provide meaningful local improvements without sacrificing global performance. Detailed quantitative results and full comparisons are provided in the Appendices~\ref{app:0.3dir} and~\ref{app:0.5dir}..

\subsection{Robustness in Pathological Settings}
\label{sec:pathological_svhn}

The results for the pathological setting, where clients hold disjoint subsets of classes, are presented in Table~\ref{tab:pathological2}. This environment serves as a stress test for federated learning algorithms. Global models like FedAvg falter here, often failing to converge because the aggregation of completely disjoint class boundaries leads to a global model that is average at everything but good at nothing.

On SVHN, FedPrism (\textbf{94.02\%}) effectively matches the gold-standard of Local-only training (94.01\%), proving that its gating mechanism successfully filters out negative transfer from disjoint clients. In contrast, FedAvg lags behind at 79.28\%, demonstrating that forced global aggregation hurts local performance when data distributions are contradictory. FedAMP also struggles to find relevant peers, achieving only 65.39\%. The breakdown of baseline performance (e.g., FedCFL at 16.66\%) further highlights the difficulty of the pathological setting and underscores FedPrism’s robustness. 

\begin{table}[h]
\centering
\caption{Results for Pathological Non-IID Setting on SVHN. Global models struggle to converge or suffer from negative transfer, while FedPrism achieves high personalization. Results are reported as mean $\pm$ standard deviation (0.5--0.9\%). }
\label{tab:pathological2}
\resizebox{0.9\columnwidth}{!}{%
\begin{tabular}{l|c|cc}
\hline
\textbf{Dataset} & \textbf{Algorithm} & \textbf{Global Acc ($\%$) } & \textbf{Local Acc ($\%$)} \\ \hline
\multirow{7}{*}{SVHN} & \textbf{FedPrism (Ours)} & 52.46 & \textbf{94.02} \\
 & Local & 17.81 & 94.01 \\
 & FedAvg & \textbf{79.94} & 79.28 \\
 & IFCA & 23.45 & 78.97 \\
 & FedAMP & 15.94 & 65.39 \\
 & FedCFL & 6.38 & 16.66 \\ 
 & FedClust & 6.72 & 19.63 \\ \hline
\end{tabular}%
}
\end{table}

The results for pathological CIFAR-100 are provided in Appendix~\ref{sec:patho_cifar100}.

\section{Ablation Study and Sensitivity Analysis (CIFAR-10, Pathological Setting)}

This section presents the ablation study of FedPrism on CIFAR-10 under the pathological non-IID setting. In this setting, each client contains data from only a small subset of classes, creating strong heterogeneity. We analyze the contribution of model components, the role of the dual-stream inference mechanism, and the sensitivity to the global weight $\alpha$.
Across experiments, local accuracy remains stable around 83–84\%. Global accuracy varies significantly depending on the configuration, especially with changes in $\alpha$.

\subsection{Component Ablation Study}

To understand the contribution of each component, we evaluate different model configurations. Table~\ref{tab:components} reports the results.

The Pure Global configuration achieves about 38–39\% global accuracy and about 84\% local accuracy. The Full FedPrism model shows very similar performance. The difference between them is small (less than 1\%), which is within normal training variation. This indicates that in this specific setting, the global backbone is the main driver of performance.

The Pure Cluster model achieves lower global accuracy (around 33\%), showing that clustering alone is not sufficient to match the global backbone.

The Pure Private model performs poorly on global accuracy (around 12\%), which is expected since it does not share information across clients. However, it still achieves above 83\% local accuracy, showing that local training alone can solve the client-specific task.

Overall, the results suggest that the global backbone is necessary for reasonable global performance, while the private component helps maintain stable local accuracy. The cluster component provides moderate contribution but does not dominate performance.

\begin{table}[h]
\caption{Impact of FedPrism Components on Accuracy (CIFAR-10, Pathological) Results are reported as mean $\pm$ standard deviation (0.5--0.9\%).}
\label{tab:components}
\centering
\resizebox{\columnwidth}{!}{%
\begin{tabular}{l|cc|cc}
\toprule
\textbf{Configuration} & \multicolumn{2}{c|}{\textbf{Global Accuracy (\%)}} & \multicolumn{2}{c}{\textbf{Local Accuracy (\%)}} \\
 & Final & Best & Final & Best \\
\midrule
\textbf{Full FedPrism} & 39.22 & 43.99 & 84.28 & 84.58 \\
Pure Global & 38.49 & 43.07 & 84.05 & 84.30 \\
Global + Cluster  & 36.46 & 43.56 & 84.07 & 84.37 \\
Pure Cluster & 33.45 & 38.67 & 83.74 & 84.10 \\
Pure Private & 11.95 & 16.10 & 83.36 & 83.57 \\
\bottomrule
\end{tabular}%
}
\end{table}

\subsection{Dual-Stream Mechanism Analysis}

We next examine the importance of the local expert during inference by varying its inference weight. A weight of 0.0 corresponds to using only the global backbone, while 1.0 corresponds to using only the local expert.

Table~\ref{tab:weight} shows that when the local expert weight is 0.0, local accuracy drops to about 12\%, which matches the global accuracy. This indicates that the global backbone alone cannot handle the pathological setting.

When the local expert is given any non-zero weight (0.2 or higher), local accuracy immediately increases to above 82–83\%. The exact differences between 0.2, 0.5, 0.8, and 1.0 are small (within about 1\%) and should not be overinterpreted. The key observation is that including the local expert is necessary for strong local performance.

\begin{table}[h]
\caption{Effect of Inference Weighting (CIFAR-10, Pathological) Results are reported as mean $\pm$ standard deviation (0.5--0.9\%).}
\label{tab:weight}
\centering
\resizebox{\columnwidth}{!}{%
\begin{tabular}{c|cc|cc}
\toprule
\textbf{Weight} & \multicolumn{2}{c|}{\textbf{Global Accuracy (\%)}} & \multicolumn{2}{c}{\textbf{Local Accuracy (\%)}} \\
 & Final & Best & Final & Best \\
\midrule
0.0 & 12.36 & 16.91 & 12.36 & 16.91 \\
0.2 & 11.17 & 17.16 & 82.38 & 83.45 \\
0.5 & 11.84 & 17.17 & 83.39 & 83.57 \\
0.8 & 11.32 & 17.07 & 83.48 & 83.48 \\
1.0 & 10.92 & 17.15 & 83.16 & 83.18 \\
\bottomrule
\end{tabular}%
}
\end{table}

\subsection{Sensitivity to Global Weight $\alpha$}

Finally, we analyze sensitivity to the global weight $\alpha$ while keeping $\beta = 0.1$ fixed. Results are shown in Table~\ref{tab:sensitivity}.

For $\alpha \le 0.5$, global accuracy remains low (around 10–12\%). This indicates that the global component is too weak to learn meaningful shared representations under strong non-IID drift.

When $\alpha$ increases to 0.7, global accuracy improves noticeably to about 22\%. At $\alpha = 0.9$, global accuracy increases further to about 38\%. This is a clear and substantial change, well beyond small training noise.

Local accuracy remains stable around 83–84\% across all $\alpha$ values, with only minor variation. Therefore, increasing $\alpha$ mainly improves global performance without harming local performance.

In this setting, a high global weight is necessary for stable global learning, while local performance is largely maintained by the private component.


An additional ablation study under Dirichlet $\alpha=0.1$ is presented in Appendix~\ref{sec:ablation_fmnist}.

\begin{table}[h]
\caption{Sensitivity to Global Weight $\alpha$ (CIFAR-10, Pathological) Results are reported as mean $\pm$ standard deviation (0.5--0.9\%).}
\label{tab:sensitivity}
\centering
\resizebox{\columnwidth}{!}{%
\begin{tabular}{cc|cc|cc}
\toprule
\textbf{$\alpha$} & \textbf{$\beta$} 
& \multicolumn{2}{c|}{\textbf{Global Accuracy (\%)}} 
& \multicolumn{2}{c}{\textbf{Local Accuracy (\%)}} \\
\cline{3-6}
& & Final & Best & Final & Best \\
\midrule
0.1 & 0.1 & 11.91 & 15.80 & 83.66 & 83.70 \\
0.3 & 0.1 & 10.32 & 17.56 & 83.45 & 83.60 \\
0.5 & 0.1 & 10.85 & 18.78 & 83.74 & 83.83 \\
0.7 & 0.1 & 22.22 & 30.99 & 83.64 & 83.69 \\
0.9 & 0.1 & 38.47 & 43.99 & 84.28 & 84.58 \\
\bottomrule
\end{tabular}%
}
\end{table}

\section{Conclusion}

This paper introduced \textbf{FedPrism}, an adaptive personalized federated learning framework designed to address severe statistical heterogeneity under non-IID data. FedPrism decomposed client models into Global, Cluster, and Private components through a structured Prism Decomposition strategy and incorporated a Dual-Stream architecture with confidence-aware routing to balance generalization and specialization.

Extensive experiments across multiple benchmarks and heterogeneity regimes demonstrated that FedPrism consistently outperformed global aggregation, clustered, and personalized baselines under moderate to extreme non-IID settings. In highly heterogeneous scenarios (Dirichlet $\alpha=0.1$ and pathological partitions), FedPrism achieved substantial improvements in local accuracy while maintaining competitive global performance. Ablation studies confirmed that the global backbone was essential for stable global learning, the private component preserved strong personalization, and the dual-stream mechanism was necessary to prevent negative transfer.

Overall, the results established FedPrism as a robust and flexible solution for federated learning in heterogeneous environments, effectively mitigating negative transfer while preserving both personalization and shared knowledge. Future work may explore scaling to larger models and extending the framework to additional modalities and dynamic real-world deployments.





\bibliographystyle{acm}
\bibliography{sample-base}

@inproceedings{mcmahan2017communication,
  title={Communication-efficient learning of deep networks from decentralized data},
  author={McMahan, Brendan and Moore, Eider and Ramage, Daniel and Hampson, Seth and y Arcas, Blaise Aguera},
  booktitle={Artificial intelligence and statistics},
  pages={1273--1282},
  year={2017},
  organization={PMLR}
}

@article{li2021survey,
  title={A survey on federated learning systems: Vision, hype and reality for data privacy and protection},
  author={Li, Qinbin and Wen, Zeyi and Wu, Zhaomin and Hu, Sixu and Wang, Naibo and Li, Yuan and Liu, Xu and He, Bingsheng},
  journal={IEEE Transactions on Knowledge and Data Engineering},
  volume={35},
  number={4},
  pages={3347--3366},
  year={2021},
  publisher={IEEE}
}

@article{li2020federated,
  title={Federated learning: Challenges, methods, and future directions},
  author={Li, Tian and Sahu, Anit Kumar and Talwalkar, Ameet and Smith, Virginia},
  journal={IEEE signal processing magazine},
  volume={37},
  number={3},
  pages={50--60},
  year={2020},
  publisher={IEEE}
}

@article{sun2022decentralized,
  title={Decentralized federated averaging},
  author={Sun, Tao and Li, Dongsheng and Wang, Bao},
  journal={IEEE Transactions on Pattern Analysis and Machine Intelligence},
  volume={45},
  number={4},
  pages={4289--4301},
  year={2022},
  publisher={IEEE}
}

@inproceedings{islam2024fedclust,
  title={Fedclust: Tackling data heterogeneity in federated learning through weight-driven client clustering},
  author={Islam, Md Sirajul and Javaherian, Simin and Xu, Fei and Yuan, Xu and Chen, Li and Tzeng, Nian-Feng},
  booktitle={Proceedings of the 53rd International Conference on Parallel Processing},
  pages={474--483},
  year={2024}
}

@article{ghosh2020efficient,
  title={An efficient framework for clustered federated learning},
  author={Ghosh, Avishek and Chung, Jichan and Yin, Dong and Ramchandran, Kannan},
  journal={Advances in neural information processing systems},
  volume={33},
  pages={19586--19597},
  year={2020}
}

@inproceedings{huang2021personalized,
  title={Personalized cross-silo federated learning on non-iid data},
  author={Huang, Yutao and Chu, Lingyang and Zhou, Zirui and Wang, Lanjun and Liu, Jiangchuan and Pei, Jian and Zhang, Yong},
  booktitle={Proceedings of the AAAI conference on artificial intelligence},
  volume={35},
  number={9},
  pages={7865--7873},
  year={2021}
}

@article{lu2024federated,
  title={Federated learning with non-iid data: A survey},
  author={Lu, Zili and Pan, Heng and Dai, Yueyue and Si, Xueming and Zhang, Yan},
  journal={IEEE Internet of Things Journal},
  volume={11},
  number={11},
  pages={19188--19209},
  year={2024},
  publisher={IEEE}
}

@article{theologitis2024communication,
  title={Communication-Efficient Distributed Deep Learning via Federated Dynamic Averaging},
  author={Theologitis, Michail and Frangias, Georgios and Anestis, Georgios and Samoladas, Vasilis and Deligiannakis, Antonios},
  journal={arXiv preprint arXiv:2405.20988},
  year={2024}
}

@inproceedings{lutz2025optimizing,
  title={Optimizing federated learning by entropy-based client selection},
  author={Lutz, Andreas and Steidl, Gabriele and M{\"u}ller, Karsten and Samek, Wojciech},
  booktitle={2025 3rd International Conference on Federated Learning Technologies and Applications (FLTA)},
  pages={218--227},
  year={2025},
  organization={IEEE}
}

@article{lecun2002gradient,
  title={Gradient-based learning applied to document recognition},
  author={LeCun, Yann and Bottou, L{\'e}on and Bengio, Yoshua and Haffner, Patrick},
  journal={Proceedings of the IEEE},
  volume={86},
  number={11},
  pages={2278--2324},
  year={2002},
  publisher={Ieee}
}

@article{sattler2020clustered,
  title={Clustered federated learning: Model-agnostic distributed multitask optimization under privacy constraints},
  author={Sattler, Felix and M{\"u}ller, Klaus-Robert and Samek, Wojciech},
  journal={IEEE transactions on neural networks and learning systems},
  volume={32},
  number={8},
  pages={3710--3722},
  year={2020},
  publisher={IEEE}
}

@inproceedings{collins2021exploiting,
  title={Exploiting shared representations for personalized federated learning},
  author={Collins, Liam and Hassani, Hamed and Mokhtari, Aryan and Shakkottai, Sanjay},
  booktitle={International conference on machine learning},
  pages={2089--2099},
  year={2021},
  organization={PMLR}
}

@article{t2020personalized,
  title={Personalized federated learning with moreau envelopes},
  author={T Dinh, Canh and Tran, Nguyen and Nguyen, Josh},
  journal={Advances in neural information processing systems},
  volume={33},
  pages={21394--21405},
  year={2020}
}

@inproceedings{briggs2020federated,
  title={Federated learning with hierarchical clustering of local updates to improve training on non-IID data},
  author={Briggs, Christopher and Fan, Zhong and Andras, Peter},
  booktitle={2020 international joint conference on neural networks (IJCNN)},
  pages={1--9},
  year={2020},
  organization={IEEE}
}

@article{smith2017federated,
  title={Federated multi-task learning},
  author={Smith, Virginia and Chiang, Chao-Kai and Sanjabi, Maziar and Talwalkar, Ameet S},
  journal={Advances in neural information processing systems},
  volume={30},
  year={2017}
}

@article{marfoq2021federated,
  title={Federated multi-task learning under a mixture of distributions},
  author={Marfoq, Othmane and Neglia, Giovanni and Bellet, Aur{\'e}lien and Kameni, Laetitia and Vidal, Richard},
  journal={Advances in neural information processing systems},
  volume={34},
  pages={15434--15447},
  year={2021}
}

@article{mansour2020three,
  title={Three approaches for personalization with applications to federated learning},
  author={Mansour, Yishay and Mohri, Mehryar and Ro, Jae and Suresh, Ananda Theertha},
  journal={arXiv preprint arXiv:2002.10619},
  year={2020}
}

@article{kendall2017uncertainties,
  title={What uncertainties do we need in bayesian deep learning for computer vision?},
  author={Kendall, Alex and Gal, Yarin},
  journal={Advances in neural information processing systems},
  volume={30},
  year={2017}
}

@article{shazeer2017outrageously,
  title={Outrageously large neural networks: The sparsely-gated mixture-of-experts layer},
  author={Shazeer, Noam and Mirhoseini, Azalia and Maziarz, Krzysztof and Davis, Andy and Le, Quoc and Hinton, Geoffrey and Dean, Jeff},
  journal={arXiv preprint arXiv:1701.06538},
  year={2017}
}

@article{jacobs1991adaptive,
  title={Adaptive mixtures of local experts},
  author={Jacobs, Robert A and Jordan, Michael I and Nowlan, Steven J and Hinton, Geoffrey E},
  journal={Neural computation},
  volume={3},
  number={1},
  pages={79--87},
  year={1991},
  publisher={MIT Press}
}

@inproceedings{yang2023impact,
  title={An impact study of concept drift in federated learning},
  author={Yang, Guanhui and Chen, Xiaoting and Zhang, Tengsen and Wang, Shuo and Yang, Yun},
  booktitle={2023 IEEE International Conference on Data Mining (ICDM)},
  pages={1457--1462},
  year={2023},
  organization={IEEE}
}

@article{krizhevsky2012imagenet,
  title={Imagenet classification with deep convolutional neural networks},
  author={Krizhevsky, Alex and Sutskever, Ilya and Hinton, Geoffrey E},
  journal={Advances in neural information processing systems},
  volume={25},
  year={2012}
}

@article{xiao2017fashion,
  title={Fashion-mnist: a novel image dataset for benchmarking machine learning algorithms},
  author={Xiao, Han and Rasul, Kashif and Vollgraf, Roland},
  journal={arXiv preprint arXiv:1708.07747},
  year={2017}
}

@article{goodfellow2013multi,
  title={Multi-digit number recognition from street view imagery using deep convolutional neural networks},
  author={Goodfellow, Ian J and Bulatov, Yaroslav and Ibarz, Julian and Arnoud, Sacha and Shet, Vinay},
  journal={arXiv preprint arXiv:1312.6082},
  year={2013}
}

@inproceedings{karimireddy2020scaffold,
  title={Scaffold: Stochastic controlled averaging for federated learning},
  author={Karimireddy, Sai Praneeth and Kale, Satyen and Mohri, Mehryar and Reddi, Sashank and Stich, Sebastian and Suresh, Ananda Theertha},
  booktitle={International conference on machine learning},
  pages={5132--5143},
  year={2020},
  organization={PMLR}
}

@inproceedings{li2021ditto,
  title={Ditto: Fair and robust federated learning through personalization},
  author={Li, Tian and Hu, Shengyuan and Beirami, Ahmad and Smith, Virginia},
  booktitle={International conference on machine learning},
  pages={6357--6368},
  year={2021},
  organization={PMLR}
}

@article{zhao2018federated,
  title={Federated learning with non-iid data},
  author={Zhao, Yue and Li, Meng and Lai, Liangzhen and Suda, Naveen and Civin, Damon and Chandra, Vikas},
  journal={arXiv preprint arXiv:1806.00582},
  year={2018}
}

@article{kairouz2021advances,
  title={Advances and open problems in federated learning},
  author={Kairouz, Peter and McMahan, H Brendan and Avent, Brendan and Bellet, Aur{\'e}lien and Bennis, Mehdi and Bhagoji, Arjun Nitin and Bonawitz, Kallista and Charles, Zachary and Cormode, Graham and Cummings, Rachel and others},
  journal={Foundations and trends{\textregistered} in machine learning},
  volume={14},
  number={1--2},
  pages={1--210},
  year={2021},
  publisher={Now Publishers, Inc.}
}

@article{yan2023clustered,
  title={Clustered federated learning in heterogeneous environment},
  author={Yan, Yihan and Tong, Xiaojun and Wang, Shen},
  journal={IEEE Transactions on Neural Networks and Learning Systems},
  volume={35},
  number={9},
  pages={12796--12809},
  year={2023},
  publisher={IEEE}
}

@article{tan2022towards,
  title={Towards personalized federated learning},
  author={Tan, Alysa Ziying and Yu, Han and Cui, Lizhen and Yang, Qiang},
  journal={IEEE transactions on neural networks and learning systems},
  volume={34},
  number={12},
  pages={9587--9603},
  year={2022},
  publisher={IEEE}
}

@article{arivazhagan2019federated,
  title={Federated learning with personalization layers},
  author={Arivazhagan, Manoj Ghuhan and Aggarwal, Vinay and Singh, Aaditya Kumar and Choudhary, Sunav},
  journal={arXiv preprint arXiv:1912.00818},
  year={2019}
}

@article{ikotun2023k,
  title={K-means clustering algorithms: A comprehensive review, variants analysis, and advances in the era of big data},
  author={Ikotun, Abiodun M and Ezugwu, Absalom E and Abualigah, Laith and Abuhaija, Belal and Heming, Jia},
  journal={Information Sciences},
  volume={622},
  pages={178--210},
  year={2023},
  publisher={Elsevier}
}

@article{kuhn1955hungarian,
  title={The Hungarian method for the assignment problem},
  author={Kuhn, Harold W},
  journal={Naval research logistics quarterly},
  volume={2},
  number={1-2},
  pages={83--97},
  year={1955},
  publisher={Wiley Online Library}
}

\clearpage
\section{Appendix}

\subsection{Performance under Moderate Heterogeneity ($\alpha=0.3$)}
\label{app:0.3dir}
As we relax the heterogeneity constraint to $\alpha=0.3$, we observe a shift in the performance landscape as shown in Table~\ref{tab:dir_0.3}. On CIFAR-10, \textbf{FedPrism} achieves a Global/Local accuracy balance of \textbf{41.87\% / 66.52\%}, whereas FedAvg sits at 46.57\% / 46.95\%. The significant boost in personalization (+20\%) with only a minor drop in global generalization highlights FedPrism's effectiveness in scenarios where one-size-fits-all models like FedAvg are suboptimal. 

For CIFAR-100, the intrinsic difficulty of the task amplifies the benefits of our approach. FedPrism reaches \textbf{23.45\%} Local Accuracy, surpassing FedAvg (15.24\%) and IFCA (13.51\%). This confirms that even with moderate data sharing potential, personalization mechanisms are essential for scalability in complex tasks with many classes.

\begin{table}[h]
\centering
\caption{Performance Comparison under Moderate Heterogeneity (Dirichlet $\alpha=0.3$). Results are reported as mean $\pm$ standard deviation (0.5--0.9\%).}
\label{tab:dir_0.3}
\resizebox{\columnwidth}{!}{%
\begin{tabular}{l|cc|cc|cc|cc}
\hline
\multirow{2}{*}{\textbf{Algorithm}} & \multicolumn{2}{c|}{\textbf{SVHN}} & \multicolumn{2}{c|}{\textbf{FMNIST}} & \multicolumn{2}{c|}{\textbf{CIFAR-10}} & \multicolumn{2}{c}{\textbf{CIFAR-100}} \\ \cline{2-9} 
 & \textit{Glob} & \textit{Loc} & \textit{Glob} & \textit{Loc} & \textit{Glob} & \textit{Loc} & \textit{Glob} & \textit{Loc} \\ \hline
\textbf{FedPrism (Ours)} & 81.73 & \textbf{84.95} & 82.59 & \textbf{90.53} & 41.87 & \textbf{66.52} & 13.64 & \textbf{23.45} \\ \hline
FedAvg & \textbf{86.94} & 86.40 & 87.10 & 86.46 & 46.57 & 46.95 & \textbf{15.31} & 15.24 \\
IFCA & 32.22 & 82.77 & 52.5 & 88.76 & 28.21 & 51.93 & 8.61 & 13.51 \\
FedClust & 78.96 & 70.39 & 78.73 & 69.32 & 32.57 & 31.16 & 4.17 & 6.30 \\
FedAMP & 22.94 & 78.48 & 32.27 & 89.83 & 16.16 & 62.09 & 3.33 & 22.22 \\
FedCFL & 19.58 & 39.01 & 14.13 & 73.03 & 10.00 & 16.02 & 1.39 & 3.83 \\
Local & 36.71 & 82.06 & 47.38 & 90.13 & 21.37 & 66.12 & 5.00 & 22.13 \\ \hline
\end{tabular}%
}
\end{table}

\subsection{Performance under Low Heterogeneity ($\alpha=0.5$)}
\label{app:0.5dir}
Table~\ref{tab:dir_0.5} presents results for the least heterogeneous setting ($\alpha=0.5$), where client distributions are more uniform. Even here, FedPrism maintains strong competitiveness. On CIFAR-100, our method achieves \textbf{18.04\%} Local Accuracy compared to Local training's 15.45\% and FedAvg's 13.88\%. This result is notable because typically, as data becomes more IID, the advantage of specialized routing diminishes; however, FedPrism continues to extract useful personalization gains without succumbing to global model collapse.

\begin{table}[h]
\centering
\caption{Performance Comparison under Low Heterogeneity (Dirichlet $\alpha=0.5$). Results are reported as mean $\pm$ standard deviation (0.5--0.9\%).}
\label{tab:dir_0.5}
\resizebox{\columnwidth}{!}{%
\begin{tabular}{l|cc|cc|cc|cc}
\hline
\multirow{2}{*}{\textbf{Algorithm}} & \multicolumn{2}{c|}{\textbf{SVHN}} & \multicolumn{2}{c|}{\textbf{FMNIST}} & \multicolumn{2}{c|}{\textbf{CIFAR-10}} & \multicolumn{2}{c}{\textbf{CIFAR-100}} \\ \cline{2-9} 
 & \textit{Glob} & \textit{Loc} & \textit{Glob} & \textit{Loc} & \textit{Glob} & \textit{Loc} & \textit{Glob} & \textit{Loc} \\ \hline
\textbf{FedPrism (Ours)} & 86.94 & \textbf{86.85} & {88.12} & 88.01 & \textbf{50.22} & 49.92 & \textbf{14.47} & \textbf{18.04} \\ \hline
FedAvg & 86.85 & 86.80 & 87.33 & 87.24 & 48.73 & 48.77 & 13.89 & 13.88 \\
IFCA & 67.63& 86.27 & 61.39 & 88.51 & 30.61 & 48.65 & 9.05 & 15.15 \\
FedClust & 79.66 & 76.23 & 85.19 & 81.09 & 22.51 & 26.66 & 5.57 & 5.45 \\
FedAMP & 33.20 & 79.49 & 48.66 & 87.21 & 18.39 & 58.58 & 3.61 & 13.58 \\
FedCFL & 29.23 & 56.85 & 75.44 & 80.55 & 10.41 & 19.95 & 1.17 & 3.87 \\
Local & 43.23 & 79.94 & 56.08 & 87.11 & 24.40 & \textbf{55.99} & 5.05 & 15.45 \\ \hline
\end{tabular}%
}
\end{table}

\subsection{Robustness in Pathological Settings}
\label{sec:patho_cifar100}

The results for the pathological setting, where clients hold disjoint subsets of classes, are presented in Table~\ref{tab:pathological}. This environment serves as a stress test for federated learning algorithms. Global models like FedAvg falter here, often failing to converge because the aggregation of completely disjoint class boundaries leads to a global model that is average at everything but good at nothing.

FedPrism, however, thrives in this environment. On CIFAR-100, it achieves a Local Accuracy of \textbf{87.60\%}, a massive improvement over FedAMP's 56.40\% and FedAvg's 5.95\%. Crucially, it outperforms the Local-only baseline (86.35\%), proving that the global backbone contributes positively even when local datasets are highly disjoint. The breakdown of baseline performance (FedCFL at 4.76\%, FedClust at 5.53\%) underscores the difficulty of the task. 

\begin{table}[h]
\centering
\caption{Results for Pathological Non-IID Setting on CIFAR-100. Global models struggle to converge or suffer from negative transfer, while FedPrism achieves high personalization. Results are reported as mean $\pm$ standard deviation (0.5--0.9\%).}
\label{tab:pathological}
\resizebox{0.9\columnwidth}{!}{%
\begin{tabular}{l|c|cc}
\hline
\textbf{Dataset} & \textbf{Algorithm} & \textbf{Global Acc} & \textbf{Local Acc} \\ \hline
\multirow{7}{*}{CIFAR-100} & \textbf{FedPrism (Ours)} & 1.04 & \textbf{87.60} \\
 & Local & 1.73 & 86.35 \\
 & FedAMP & 1.00 & 56.40 \\
 & IFCA & 2.48 & 13.04 \\
 & FedAvg & 5.95 & 5.95 \\ 
 & FedClust & 1.00 & 5.53 \\
 & FedCFL & 1.00 & 4.76 \\ \hline
\end{tabular}%
}
\end{table}

\subsection{Ablation Study: FMNIST (Dirichlet $\alpha=0.1$)}
\label{sec:ablation_fmnist}

This section presents an ablation and sensitivity analysis for the Fashion-MNIST (FMNIST) dataset under a non-IID setting generated via Dirichlet distribution with $\alpha=0.1$. This partitioning simulates high statistical heterogeneity among clients. Results are reported as mean $\pm$ standard deviation (0.5--0.9\%).

\subsection{Component Ablation Study}
We evaluated the contribution of individual FedPrism components: 'Pure Global' (backbone only), 'Pure Cluster' (prototypes only), 'Pure Private' (private headers only), and 'Global+Cluster' (no private component). The results are summarized in Table \ref{tab:fmnist_ablation}.

\begin{table}[h]
\centering
\caption{Impact of Components (FMNIST, Dirichlet $\alpha=0.1$)}
\label{tab:fmnist_ablation}
\resizebox{\columnwidth}{!}{%
\begin{tabular}{lcc}
\toprule
\textbf{Configuration} & \textbf{Final Global Acc (\%)} & \textbf{Final Local Acc (\%)} \\
\midrule
Pure Global & \textbf{84.81} & \textbf{95.70} \\
Pure Cluster & 82.37 & 95.54 \\
Global + Cluster (No Private) & 84.21 & 95.51 \\
Pure Private & 19.98 & 95.15 \\
\bottomrule
\end{tabular}%
}
\end{table}

The data indicates effectively high performance across most configurations involving shared knowledge. The 'Pure Global' model achieves the highest global accuracy (84.81\%) and local accuracy (95.70\%), suggesting that for FMNIST, a strong central backbone is sufficient to capture general features even under heterogeneity. The 'Pure Cluster' approach also performs robustly (82.37\%), confirming that prototype-based clustering can effectively group similar clients. Notably, the 'Pure Private' model completely fails to generalize globally ($\sim 20\%$) but maintains high local accuracy ($\sim 95\%$), reinforcing the role of local experts in solving client-specific tasks.

\subsection{Dual Mechanism Analysis}
We analyzed the effect of the inference weight in the Dual-Stream mechanism, where a weight of 0.0 implies reliance solely on the global backbone and 1.0 solely on the local expert.

\begin{table}[h]
\centering
\caption{Effect of Inference Weighting (FMNIST)}
\label{tab:fmnist_weight}
\resizebox{\columnwidth}{!}{%
\begin{tabular}{ccc}
\toprule
\textbf{Local Expert Weight} & \textbf{Global Acc (\%)} & \textbf{Local Acc (\%)} \\
\midrule
0.0 (Global Only) & 39.54 & 43.07 \\
0.2 & 40.09 & 93.48 \\
0.5 & 40.49 & 95.38 \\
0.8 & 40.30 & 95.18 \\
1.0 (Local Only) & 39.10 & 95.25 \\
\bottomrule
\end{tabular}%
}
\end{table}

Reference Table \ref{tab:fmnist_weight} shows a clear dependency on the local expert. When the weight is set to 0.0, local accuracy drops significantly to 43.07\%. Introducing even a small weight (0.2) restores local accuracy to $>93\%$, highlighting that personalization is essential for handling the non-IID distribution in FMNIST.

\subsection{Hyperparameter Sensitivity ($\alpha$)}
We examined the sensitivity to the global model weight $\alpha$, which balances the global backbone against cluster/private components.

\begin{table}[h]
\centering
\caption{Sensitivity to Global Weight $\alpha$ (FMNIST)}
\label{tab:fmnist_alpha}
\resizebox{\columnwidth}{!}{%
\begin{tabular}{ccc}
\toprule
\textbf{Alpha ($\alpha$)} & \textbf{Global Acc (\%)} & \textbf{Local Acc (\%)} \\
\midrule
0.1 & 27.46 & 95.19 \\
0.3 & 34.95 & 95.29 \\
0.5 & 41.83 & 95.20 \\
0.7 & 63.62 & 95.44 \\
\textbf{0.9} & \textbf{84.83} & \textbf{95.55} \\
\bottomrule
\end{tabular}%
}
\end{table}

Table \ref{tab:fmnist_alpha} reveals a strong positive correlation between $\alpha$ and global accuracy. Performance increases monotonically, peaking at $\alpha=0.9$ with 84.83\% global accuracy. This demonstrates that prioritizing the global backbone allows the model to learn a generalized representation that benefits all clients, without compromising local accuracy, which remains stable at $\sim 95\%$ across all settings.

\subsection{Data Distribution Plots}
\label{appendix:data_distribution}
\FloatBarrier
This section visualizes the non-IID data distribution across clients for 
\textbf{SVHN} and \textbf{FMNIST}. We present the class distribution for 
the first 20 clients under Dirichlet sampling ($\alpha = 0.1$) and 
Pathological partitioning.


\begin{figure}[!t]
    \centering
    \includegraphics[width=\columnwidth]{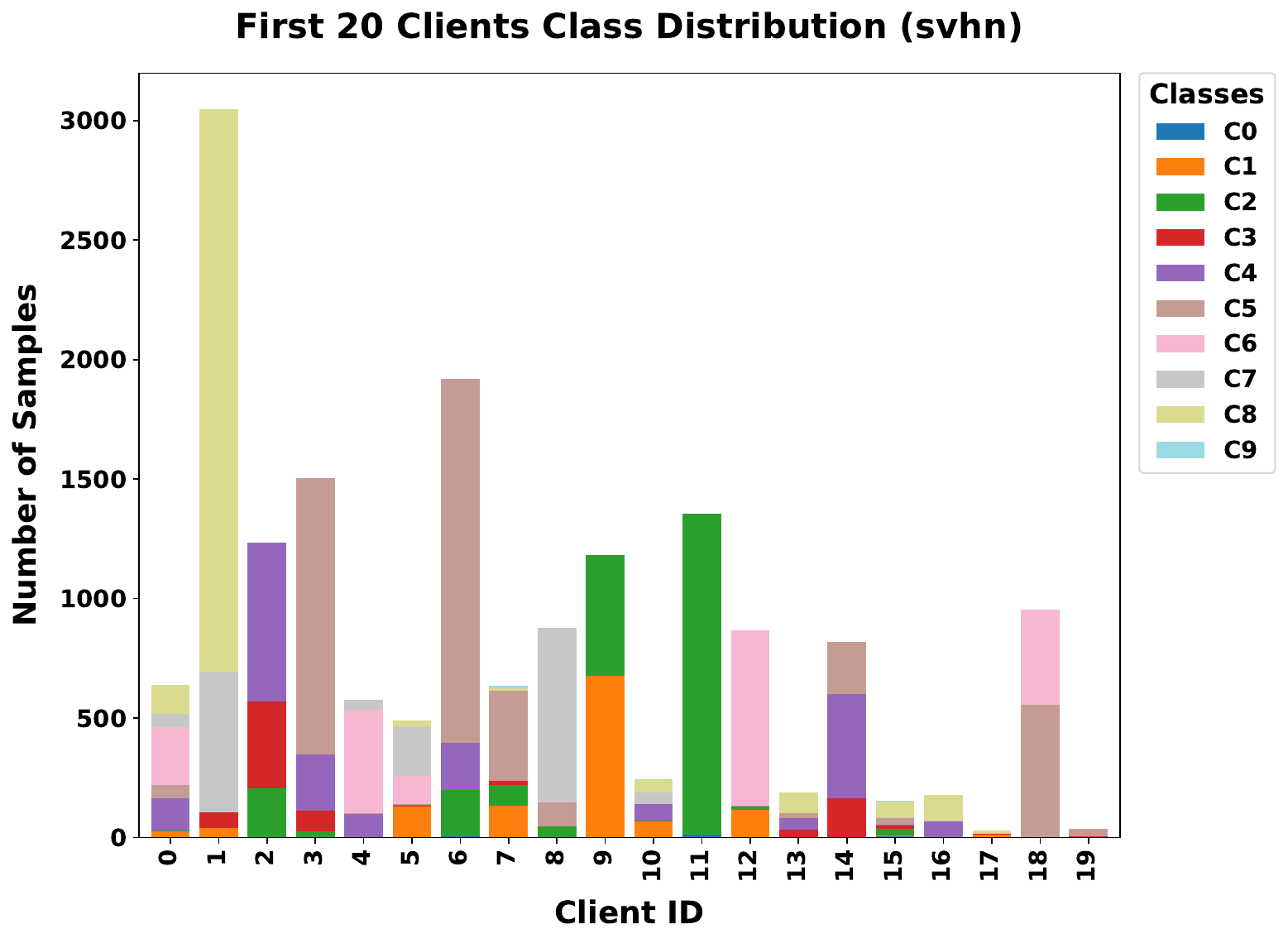}
    \Description{Class distribution for the first 20 clients on SVHN using Dirichlet sampling with alpha equal to 0.1.}
    \caption{SVHN under Dirichlet partitioning ($\alpha = 0.1$).}
    \label{fig:svhn_dirichlet}
\end{figure}

\begin{figure}[!t]
    \centering
    \includegraphics[width=\columnwidth]{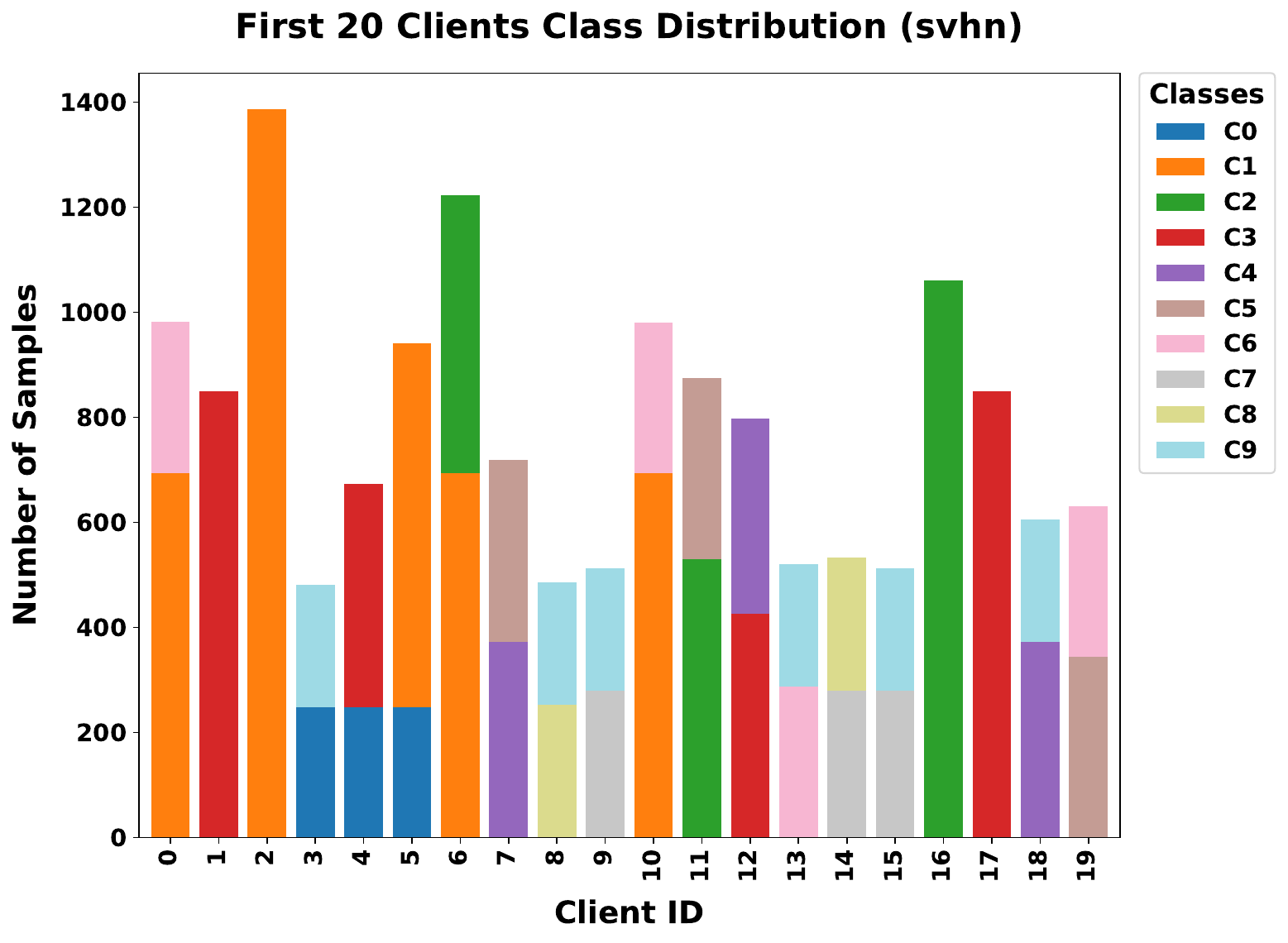}
    \Description{Class distribution for the first 20 clients on SVHN using pathological partitioning.}
    \caption{SVHN under pathological partitioning.}
    \label{fig:svhn_pathological}
\end{figure}


\begin{figure}[H]
    \centering
    \includegraphics[width=\columnwidth]{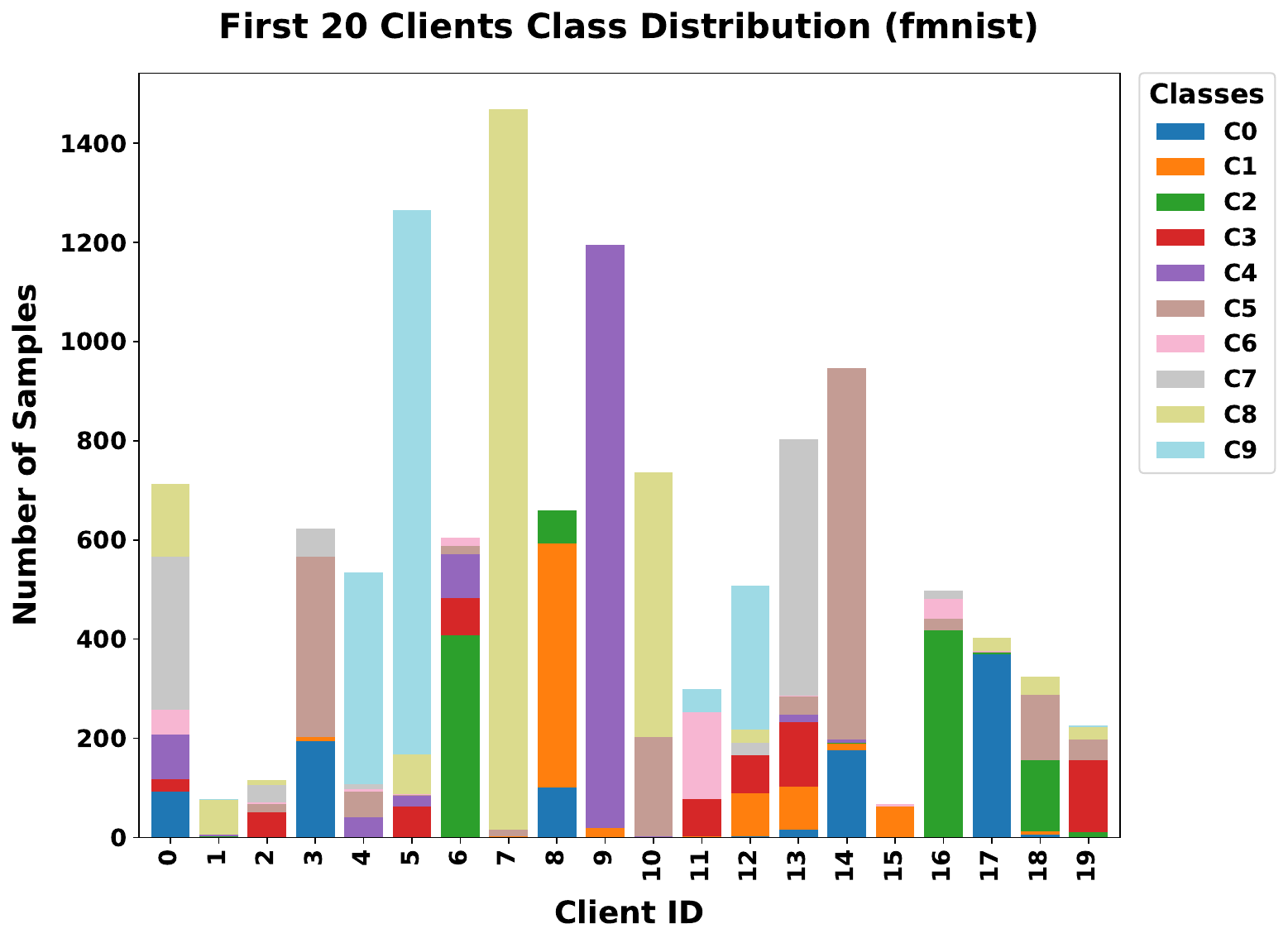}
    \Description{Class distribution for the first 20 clients on FMNIST using Dirichlet sampling with alpha equal to 0.1.}
    \caption{FMNIST under Dirichlet partitioning ($\alpha = 0.1$).}
    \label{fig:fmnist_dirichlet}
\end{figure}

\begin{figure}[H]
    \centering
    \includegraphics[width=\columnwidth]{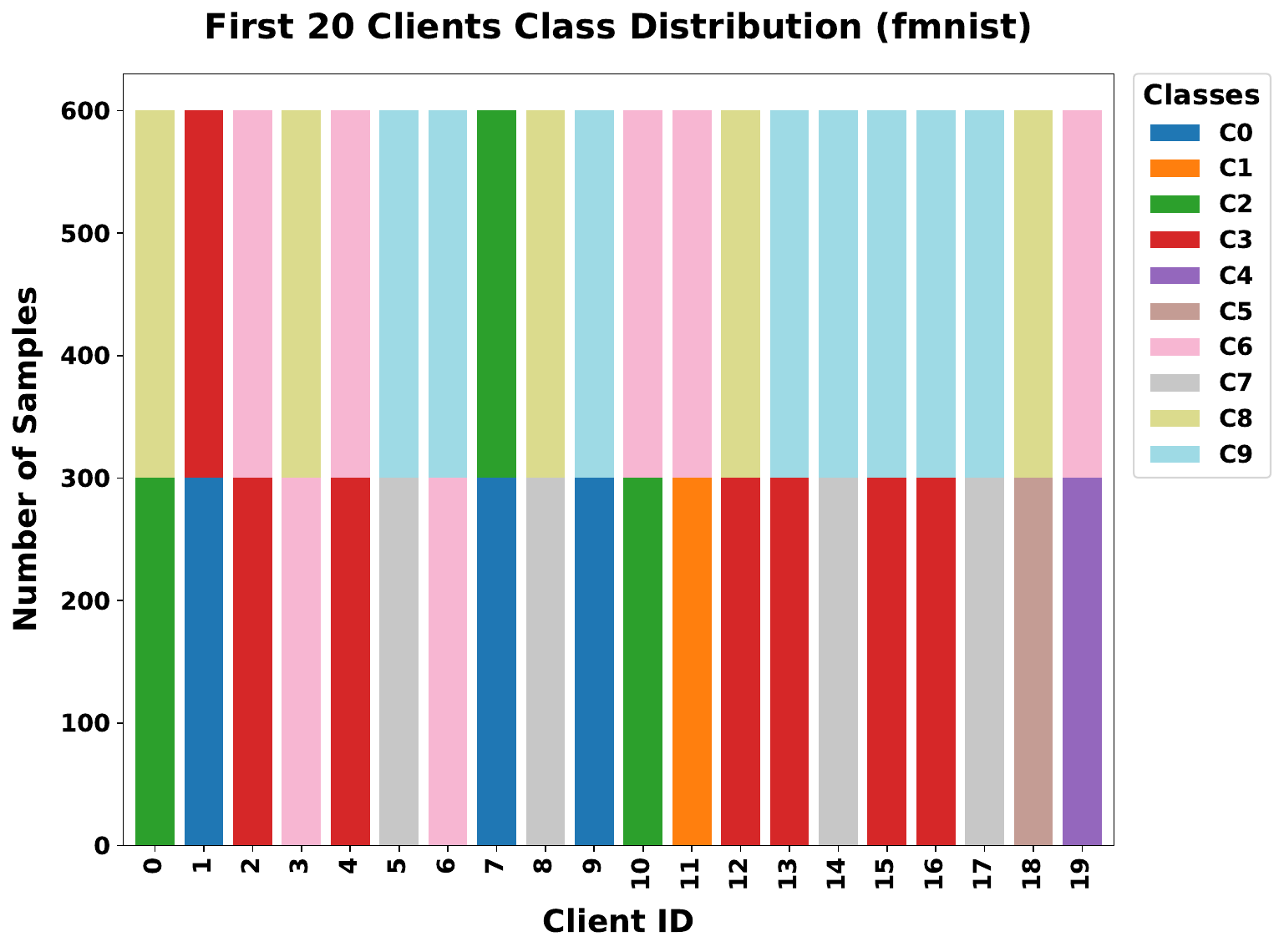}
    \Description{Class distribution for the first 20 clients on FMNIST using pathological partitioning.}
    \caption{FMNIST under pathological partitioning.}
    \label{fig:fmnist_pathological}
\end{figure}
\FloatBarrier


\end{document}